\documentclass[conference]{IEEEtran}

\PassOptionsToPackage{no-math}{fontspec}
\usepackage{xltxtra}
\AtBeginDocument{%
  \let\scshape\relax
  \renewcommand\textsc[1]{#1}%
}

\usepackage{amsmath}
\usepackage{amssymb}
\usepackage{graphicx}
\usepackage{booktabs}
\usepackage[backend=biber,style=numeric-comp,sorting=none,maxbibnames=99]{biblatex}
\usepackage[above]{placeins}
\usepackage{makecell}
\usepackage{adjustbox}
\usepackage[table]{xcolor}
\usepackage[hidelinks]{hyperref}
\hypersetup{
  pdftitle={WASABI: Whole-graph Assignment-based Stabilizer for lAne topology By Inter-frame tracking},
  pdfauthor={Tetsuhiro Uchida, Myu Sasaki, Kensho Nakajima, Yasuhiro Shimada, Toru Saito},
  pdfsubject={Lane Topology Stabilization for Autonomous Driving},
  pdfkeywords={lane topology, segment tracking with connectivity, lane connectivity (LCLC), data association, real-time post-processing, autonomous driving},
  pdfcreator={},pdfproducer={}
}

\begin{document}

\title{{\fontsize{20}{24}\selectfont WASABI: Whole-graph Assignment-based Stabilizer\\
for lAne topology By Inter-frame tracking}}

\author{%
  \IEEEauthorblockN{%
    Tetsuhiro Uchida\IEEEauthorrefmark{1},
    Myu Sasaki\IEEEauthorrefmark{1},
    Kensho Nakajima\IEEEauthorrefmark{1},
    Yasuhiro Shimada\IEEEauthorrefmark{1},
    Toru Saito\IEEEauthorrefmark{1}}
  \IEEEauthorblockA{\IEEEauthorrefmark{1}Sony Honda Mobility Inc.}
}

\maketitle

\begin{abstract}
Autonomous driving requires understanding the road as a graph of drivable
lanes and their connectivity, beyond the ego lane alone, to follow routes
through intersections and reason about cross- and merging-traffic.
Recent perception models infer such lane topology, i.e., lane
segments together with their inter-lane connectivity (LCLC), from onboard
sensors over a 360-degree BEV view.
Due to neural perception's imperfections, their outputs
retain structural instabilities such as missed detections, lost or
incorrect LCLC, over-detection, and label flicker.
This paper presents \textbf{WASABI}, a real-time post-processing pipeline
that stabilizes lane topology outputs both within and across frames by
treating lane segments and their LCLC connectivity as joint tracking
targets, under onboard real-time constraints (10~Hz / 20~ms / up to 200
input lanes).
The pipeline integrates segment tracking with connectivity, noise-robust
topology-aware refinement, and a resource-constrained real-time design.
On internal validation data (16 sequences), WASABI improves LCLC
detection F1 from 0.834 to 0.948 (+0.114, +13.6\%) and reduces centerline
lateral error from 2.50~m to 0.95~m, while reducing detection
false-positives by 24.6\%.
Temporal-stability metrics on the same data show LCLC toggle rate
reduced by 63.3\% and boundary-label flicker rate by 30.2\%, confirming
across-frame stabilization beyond per-frame accuracy.
\end{abstract}


\section{Introduction}

Autonomous driving has to plan and control the vehicle in scenes that go
beyond the single ego lane, such as following upcoming routes through
intersections or reasoning about cross- and merging-traffic from neighboring
lanes.
These operations require representing the road as a graph of drivable lanes
together with the connectivity between them.
Such a representation is captured by lane topology, i.e., the set of
lane segments together with their inter-lane connectivity
(Lane Center to Lane Center; LCLC), where LCLC describes which lane
segment is reachable from which.

\begin{figure}[!t]
  \centering
  \includegraphics[width=\columnwidth]{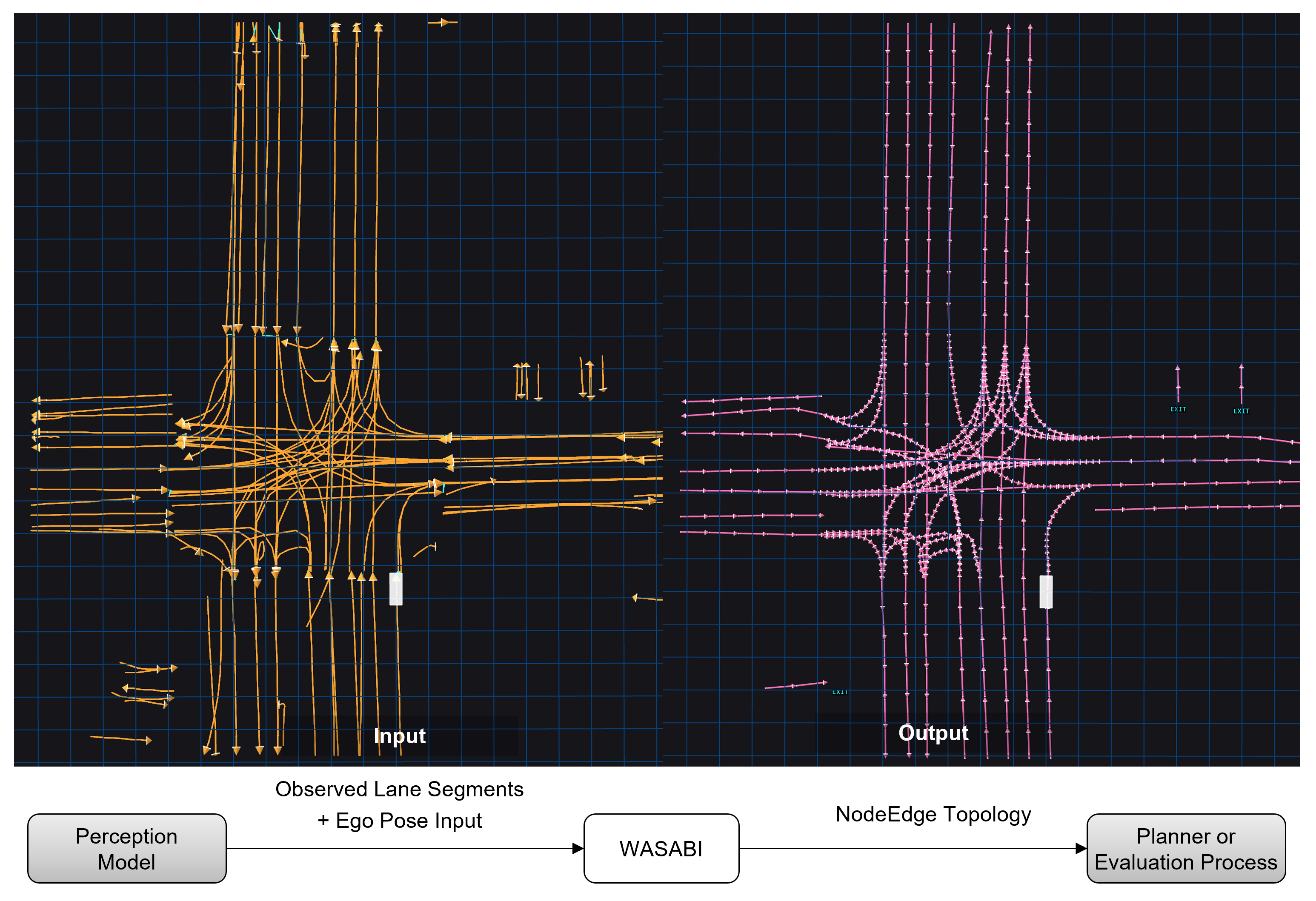}
  \caption{WASABI input and output.
    Left: noisy lane topology and ego-pose. 
    Right: stabilized NodeEdgeTopology after inter-frame tracking and post-processing.
    }
  \label{fig:overview}
\end{figure}

Recent perception models infer this representation from onboard sensors over
a 360-degree BEV
view~\cite{liao2023maptr,li2024lanesegnet,li2023toponet,wang2023openlanev2}.
Reflecting imperfections inherent to neural perception, the outputs exhibit
structural instabilities at the per-frame level. For the in-house perception
model considered in this work, the following instabilities surfaced as
downstream concerns:
(i) \textit{lane loss (missed detections of present lane segments)},
(ii) \textit{lost or incorrect LCLC},
(iii) \textit{over-detection of the same road segment}, and
(iv) \textit{flickering of boundary labels}.
These instabilities directly destabilize downstream planning and control,
manifesting as abrupt stops on ego-lane loss, route-tracking disruption from
incorrect connections, and lateral-control oscillation from node
lateral-position error.

Combining a perception model with a downstream post-processing stage to absorb
such per-frame instability is a long-standing recipe in production perception
stacks.
Multi-object tracking has been the standard recipe for object
detection~\cite{bewley2016sort,wojke2017deepsort,zhang2022bytetrack,cao2023ocsort},
but it targets axis-aligned bounding boxes representing individual objects,
whereas lane topology consists of curvilinear segments together with their
inter-segment connectivity.
This work takes the position of treating lane segments and their LCLC
connectivity jointly as tracking targets under onboard real-time constraints.

We therefore propose \textbf{WASABI}, a real-time post-processing pipeline
that stabilizes lane topology outputs from perception models both within
and across frames~(Fig.~\ref{fig:overview}).
Treating lane segments \emph{and} their inter-segment connectivity as joint
tracking targets, and operating under embedded compute budgets, are the
central design choices.
The technical contributions of this paper are as follows:

\begin{enumerate}
  \item \textbf{Segment Tracking with Connectivity}: Temporal stabilization
        that jointly tracks lane segments and their connection topology (LCLC)
        as unified tracking targets with temporal history integration.
  \item \textbf{Noise-Robust Topology-Aware Refinement}: Robust preprocessing
        and output correction that maintains road structural consistency under
        noise.
  \item \textbf{Resource-Constrained Real-Time Design}: Satisfies constraints
        of 10~Hz / 20~ms / up to 200 inputs through early-termination fallback
        and computational efficiency improvements.
\end{enumerate}

Validation on an internal evaluation dataset with a proprietary perception
model confirms improvements in lane detection and LCLC detection accuracy along
with a substantial reduction in lateral error.

\section{Related Work}

\subsection{Lane Topology Perception}

Online vectorized HD map construction has progressed from rasterized
segmentation to end-to-end lane topology reasoning.
MapTR~\cite{liao2023maptr} formulates map elements as sets of ordered polyline
points and uses a transformer decoder to produce vectorized lane boundaries.
LaneSegNet~\cite{li2024lanesegnet} extends this paradigm by representing lanes
as lane segments with associated attributes, supporting richer topological
descriptions.
TopoNet~\cite{li2023toponet} introduces explicit graph-based topology reasoning,
modeling LCLC (lane-to-lane) and LCTE (Lane Center to Traffic Element)
relationships through a dedicated graph neural network.
OpenLane-V2~\cite{wang2023openlanev2} establishes the standard benchmark for these methods.
These representative methods operate on each frame independently.
Extensions that incorporate temporal context inside the perception model have
also been proposed; StreamMapNet~\cite{yuan2024streammapnet} propagates BEV
features and detection queries across consecutive frames to produce
temporally aggregated HD map predictions.

\subsection{Multi-Object Tracking}

Temporal consistency in detection outputs is a well-studied problem in the
multi-object tracking (MOT) literature.
SORT~\cite{bewley2016sort} combines a Kalman filter for state prediction with
the Hungarian algorithm for data association, achieving real-time performance
with a simple design.
DeepSORT~\cite{wojke2017deepsort} augments SORT with a deep appearance
descriptor, reducing identity switches at the cost of additional computation.
ByteTrack~\cite{zhang2022bytetrack} improves recall via a two-stage matching scheme 
using low-confidence detections, 
while OC-SORT~\cite{cao2023ocsort} re-designs association to be observation-centric 
to address estimation errors during track re-initialization after occlusion.

These methods are designed for axis-aligned bounding boxes and rely on
appearance features or IoU overlap as the association metric, neither of which
transfers directly to curvilinear lane segments.
Furthermore, MOT tracks individual objects and does not address the temporal
consistency of inter-object connectivity.

\subsection{Curve Matching and Assignment}

Matching lane segments requires both shape similarity evaluation between curves
and global assignment.
The Fr\'echet distance~\cite{alt1995frechet} is the canonical metric for
comparing curves while respecting their parameterization, but the exact
discrete Fr\'echet distance between two $n$-point curves requires $O(n^2)$
time via dynamic programming.
For global assignment, the Hungarian algorithm~\cite{kuhn1955hungarian}
provides the optimal solution in polynomial time and is the standard choice
for MOT data association.

\subsection{Position of This Work}

Lane topology perception, whether per-frame or temporal in design, can leave
structural instabilities at the per-frame level in its outputs.
MOT provides global assignment and track state management as a post-processing
recipe but is designed for axis-aligned bounding boxes representing individual
objects, with no provision for inter-object connectivity as a tracking state.
Curve matching and assignment provide building blocks for comparing curvilinear targets, 
but no comparable post-processing pipeline integrates them into a MOT-style framework 
that jointly stabilizes inter-segment connectivity (LCLC) 
on curvilinear lane segments under onboard real-time constraints.

\section{Method}

\subsection{System Overview}

\begin{figure}[!t]
  \centering
  \includegraphics[width=\columnwidth]{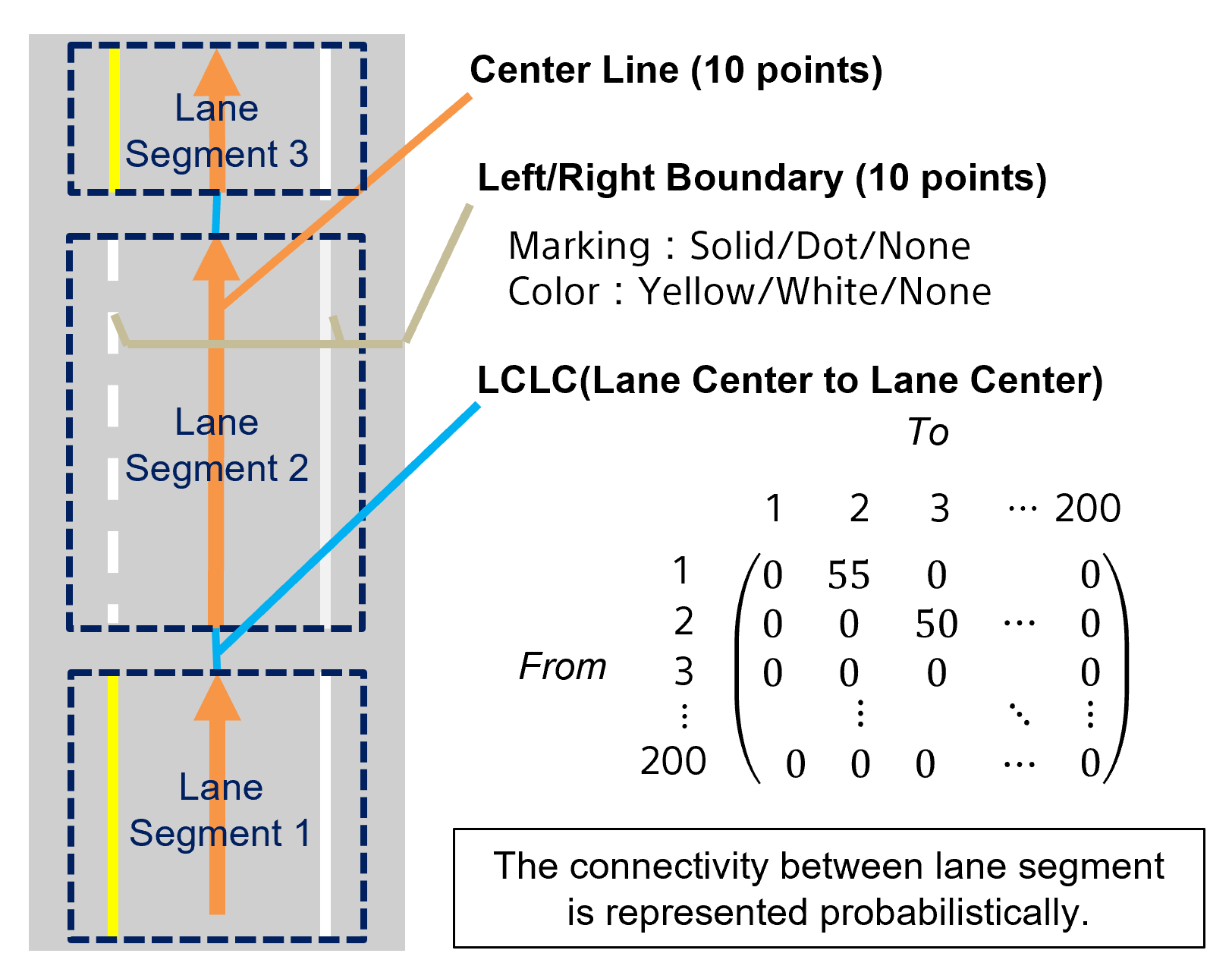}
  \caption{Conceptual diagram of lane topology. 
    Left: lane segments with geometric and semantic attributes. 
    Right: probabilistic representation of inter-segment connectivity (LCLC).}
  \label{fig:lane-topology-concept}
\end{figure}

The role of WASABI is to stabilize the lane topology output from the
perception model both within and across frames, and to deliver it to the
downstream planner in a form it can directly use.
Its inputs are the lane topology from the perception model (up to 200 lane
segments and up to $200 \times 200 = 40{,}000$ LCLC candidate connections;
see Fig.~\ref{fig:lane-topology-concept} for the underlying representation)
and ego-vehicle pose (delta pose and world pose).
Its output is NodeEdgeTopology for the downstream planner
(Fig.~\ref{fig:overview}).

\begin{figure*}[t!]
  \centering
  \includegraphics[width=\textwidth]{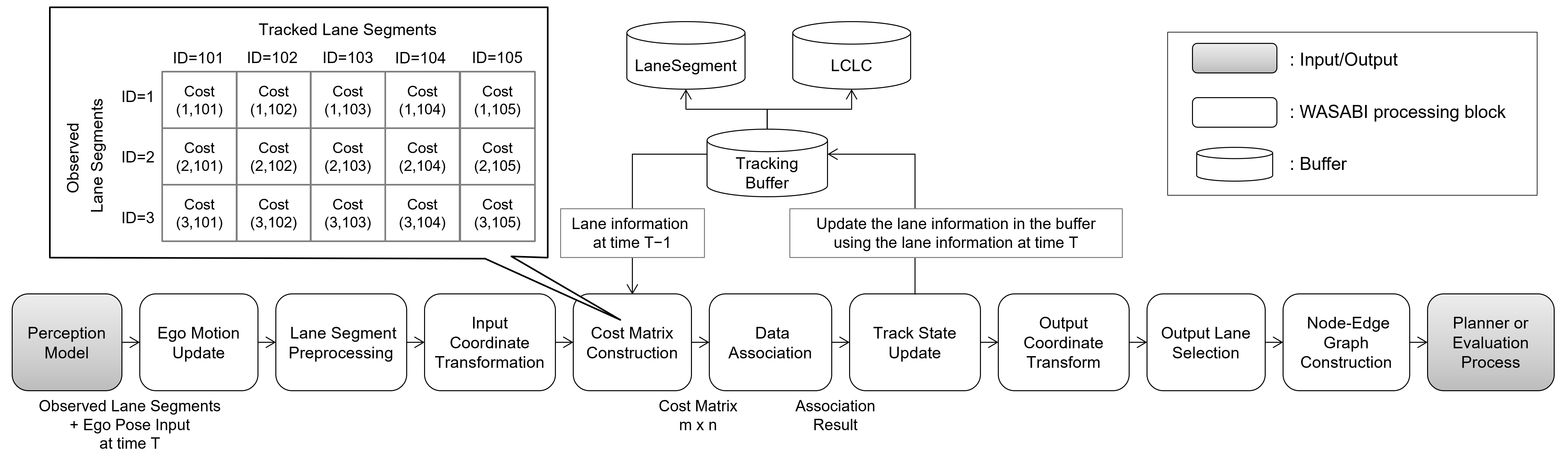}
  \caption{WASABI processing pipeline.}
  \label{fig:pipeline}
\end{figure*}

The processing pipeline is shown in Fig.~\ref{fig:pipeline}.
After ego-motion compensation via odometry update, the input lane topology is
processed in the following order:
(i) \textit{frame-internal preprocessing for single-frame noise suppression
(Section~III-C)},
(ii) \textit{cost computation, association, and state update via the tracking
buffer (Section~III-B)},
(iii) \textit{output lane selection, topology correction, and endpoint smoothing
after conversion to the ego coordinate frame (Section~III-C)}, and
(iv) \textit{conversion to NodeEdgeTopology}.

Note that the following sections describe each contribution in isolation;
the section order does not follow the execution order in Fig.~\ref{fig:pipeline}.

\subsection{Segment Tracking with Connectivity}\label{ssec:segment_tracking}

WASABI stabilizes not only the positional tracking of lane segments but also
the temporal consistency of LCLC simultaneously.
This corresponds to cost matrix construction through state update in Fig.~\ref{fig:pipeline} 
and resolves single-frame structural ambiguities via temporal integration.
In the tracking buffer, lane segments and LCLC are managed as separate
entities: each lane segment is assigned a unique Track~ID, and each LCLC is
identified by a Track~ID pair (FromID, ToID).
From up to $200 \times 200$ LCLC candidates, 
only connections passing consistency filters are retained, 
with up to $K$ stored in the buffer.

\subsubsection{Frame-to-Frame Association}\label{ssec:association}

The correspondence between observed lane segments at time $T$ (index: $k$) and 
buffered segments (index: $l$) is formulated as an assignment problem. 
Invalid pairs are gated using lane attributes, heading, lateral distance $d_{\mathrm{PtoL}}$, 
and overlap ratio, and costs are computed for the remaining candidates.
Under straight driving, the cost is defined as

\begin{equation}
\mathrm{Cost}(k, l) = d_{\mathrm{Frechet}}(k, l) + d_{\mathrm{PtoL}}(k, l)
\end{equation}

while under large rotational motion 
($\|\dot{\psi}\| > \tau_{\dot{\psi}}$), it switches to  the Overlap
Fr\'echet distance:

\begin{equation}
\mathrm{Cost}(k, l) = d_{\mathrm{OverlapFrechet}}(k, l) + d_{\mathrm{PtoL}}(k, l)
\end{equation}

To reduce computational complexity, 
both observed and buffered segments are partitioned into four heading groups, 
and assignment is solved independently using the Hungarian algorithm~\cite{kuhn1955hungarian}, 
without sacrificing association quality since correspondences across opposing 
headings are spurious by construction.

Here, $d_{\mathrm{PtoL}}$ is defined as $\max(d_1, d_2)$, 
where $d_1$ and $d_2$ are the shortest distances 
from the endpoints of the observed segment to the buffered polyline.

For shape similarity, the discrete Fr\'echet distance~\cite{alt1995frechet} is 
approximated by same-index correspondence.
Let the observed and buffered segment point sequences be $P = \{p_i\}$ and $B = \{b_i\}$:

\begin{equation}
d_{\mathrm{Frechet}}(B, P) \approx \max_i \| b_i - p_i \|
\end{equation}

reducing complexity from $O(n^2)$ to $O(n)$.
Gating restricts candidates to spatially proximate pairs, 
for which the optimal Fréchet correspondence lies near the diagonal. 
Since same-index correspondence is monotone, 
its maximum distance upper-bounds the true Fréchet distance, 
so off-diagonal pairs are overestimated and rejected rather than mis-associated. 
This validity is confirmed by the ablation study in Section IV-C.

$d_{\mathrm{OverlapFrechet}}$ evaluates the discrete Fréchet distance 
over the overlapping region, defined as the maximum matched-point distance. 
This avoids overestimated costs from non-overlapping segments caused
by lane-length variation as additional lane portions become visible and 
by endpoint drift during turning, while preserving shape similarity.

\subsubsection{State Management}

Each lane segment is tracked through four states (Tentative, Tracking,
TemporaryLost, Delete).
A new observation enters Tentative and is promoted to Tracking after
$N_{TrackStMove}$ consecutive associations; a Tracking track whose
association fails moves to TemporaryLost, returning to Tracking on
re-observation.
Deletion from TemporaryLost is governed by an observation-adaptive miss
budget $N_{\mathrm{lost}}(d)$ that grows with the detection history, so
stably detected lanes are retained longer.
Full parameter values are listed in Table~\ref{tab:params}.

\subsubsection{LCLC Temporal Consistency}

\begin{figure}[!t]
  \centering
  \includegraphics[width=\columnwidth]{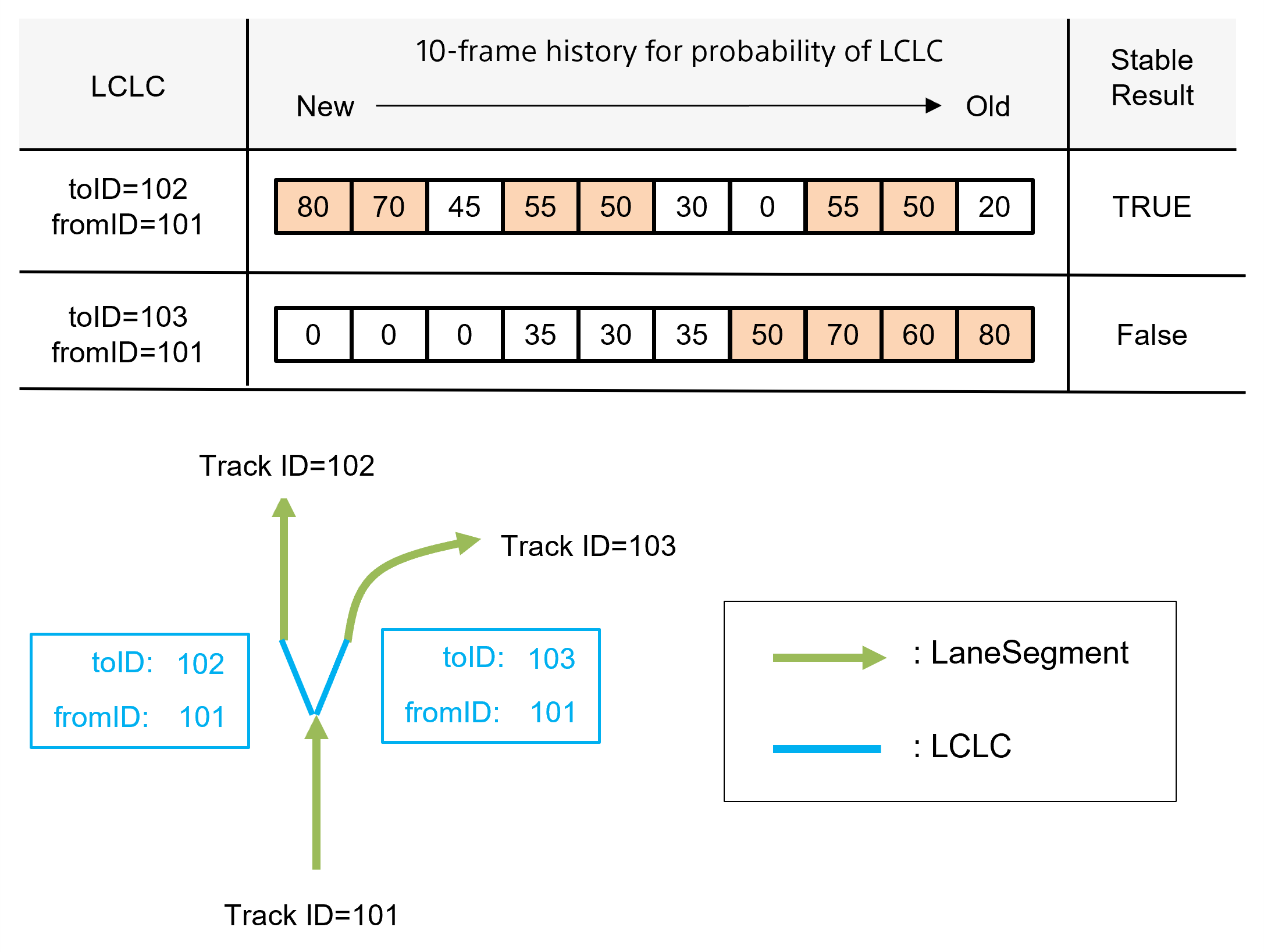}
  \caption{LCLC temporal stabilization via sliding-window majority vote. 
    Top: per-LCLC scores over a sliding window (W = 10) used for majority voting. 
    Bottom: an example of branching connections from TrackID=101 to 102 and 103.}
  \label{fig:lclcwindow}
\end{figure}

To prevent connectivity gaps in NodeEdgeTopology output, inference confidence
is integrated over history for each LCLC (Fig.~\ref{fig:lclcwindow}).
For each LCLC (FromID, ToID), the per-frame inference score $s_t$ is maintained
in a sliding window of length $W = 10$.
If the number of frames within the window satisfying $s_t \geq \tau$ reaches
$M$ or more, that LCLC is judged as valid (TRUE).
This majority-vote history integration suppresses connectivity gaps caused by
isolated false detections or temporary score drops.

\subsection{Noise-Robust Topology-Aware Refinement}\label{ssec:refinement}

The perception model output often contains duplicate lane segments and spurious LCLC connections, 
which destabilize association when directly used for tracking. 
To address this, noise-robust structuring is applied to both the input (frame-internal preprocessing) 
and output (lane selection, topology correction, and endpoint smoothing).

The major structural hyperparameters are summarized in Table I, 
with thresholds determined empirically from validation data. 
Geometric and score thresholds
($\tau_{\theta}, \tau_{L2}, \tau_{PtoL}, \tau_{ov}, \tau_{end}, \tau_{\cos}, \tau_{\dot\psi}$)
are omitted here as they are tuned per sensor configuration and dataset.

\begin{table}[t]
\caption{Implementation Parameters (Major Structural Hyperparameters)}
\label{tab:params}
\centering
\footnotesize
\begin{tabular}{lp{3.4cm}p{2.6cm}}
\hline
Symbol & Meaning & Value \\
\hline
$W$ & Sliding window length for LCLC history & 10 \\
$M$ & Majority-vote count for LCLC validity & 5 \\
$K$ & Maximum number of LCLCs held in buffer & 700 \\
$N_{TrackStMove}$ & Consecutive associations for Tentative$\to$Tracking & 3 \\
$N_{TentDrop}$ & Consecutive misses for Tentative$\to$Delete & 1 \\
$N_{promote}$ & Detection counter threshold for Confirmed promotion & 15 \\
$N_{lost}(d)$ & Allowed missed detections in TemporaryLost & $=$ detection\_counter \\
$d$ & B\'{e}zier curve degree for endpoint smoothing & Up to $N$ (FrontSide/RearSide) or $2N-1$ (BothSide) \\
$N$ & Local point sequence length for endpoint smoothing & 3 \\
\hline
\end{tabular}
\end{table}

\vspace{0.3em}

\subsubsection{Frame-Internal Preprocessing (Input Side)}\leavevmode\\

\vspace{-0.5em}

\textbf{Deduplication.} Low-confidence segments are first removed by score threshold.
Two segments $s_i, s_j$ are then judged as duplicates by the conjunction of
heading difference $\Delta\theta$, representative-point distance $d_{L2}$,
point-to-line distance $d_{\mathrm{PtoL}}$, overlap ratio $\rho$, and
attribute match $\mathrm{attr}$:

\begin{equation}
\begin{aligned}
\mathrm{dup}(i,j)
= \mathbb{I}\Big[
& \Delta\theta(s_i, s_j) \le \tau_\theta \\
& \wedge\ d_{L2}(s_i, s_j) \le \tau_{L2} \\
& \wedge\ d_{\mathrm{PtoL}}(s_i, s_j) \le \tau_{\mathrm{PtoL}} \\
& \wedge\ \rho(s_i, s_j) \ge \tau_{\mathrm{ov}} \\
& \wedge\ \mathrm{attr}(s_i) = \mathrm{attr}(s_j)
\Big]
\end{aligned}
\end{equation}

Within each duplicate group, only the representative with the highest
confidence and geometric quality is retained.

\textbf{Frame-Internal LCLC Refinement.}
To recover frequently missing connections, LCLC entries are added for segment
pairs $(u, v)$ satisfying endpoint proximity and heading alignment:

\begin{equation}
\mathrm{add}(u,v)
=
\mathbb{I}\Big[
d_{\mathrm{end}}(u,v) \le \tau_{\mathrm{end}}
\wedge \cos\phi(u,v) \ge \tau_{\mathrm{cos}}
\Big]
\end{equation}

Excessively long connections and geometrically inconsistent connections
(e.g., crossing lane boundaries) are removed, and when redundant connections
exist for the same (From, To) pair, only the representative is retained.

This preprocessing recovers missing connections while suppressing spurious ones,
stabilizing subsequent association and state update.

\vspace{0.3em}

\subsubsection{Output Selection and Correction (Output Side)}\leavevmode\\

\vspace{-0.3em}

\textbf{Output Lane Selection.}
From Tracking and TemporaryLost lanes in the buffer, 
candidates consistent with the ego travel direction and spatial proximity are selected.

Each candidate $c$ is scored by a weighted sum of three components, each normalized to $[0,1]$: 
a length score $S_{\mathrm{len}}$ (favoring longer forward extent, including connected lanes), 
a heading score $S_{\mathrm{yaw}}$ (alignment with ego heading), and 
a lateral score $S_{\mathrm{lat}}$ (proximity to the ego):
\begin{equation}
s(c) = w_{\mathrm{len}}\, S_{\mathrm{len}}(c) + w_{\mathrm{yaw}}\, S_{\mathrm{yaw}}(c) 
      + w_{\mathrm{lat}}\, S_{\mathrm{lat}}(c)
\end{equation}
The lane $c^* = \arg\max_c s(c)$ is selected as the reference lane of the ego zone.
The remaining candidates are assigned to the ego, left, and right zones relative to 
$c^*$. One priority lane is then selected per zone using the same weighted score, 
retaining the ego lane and its lateral neighbors in the output.
TemporaryLost lanes are included as candidates to prevent output gaps caused by temporary false negatives.
To limit false positives, only per-zone priority lanes are selected from the Tracking and TemporaryLost candidates.
The weights $w_{\mathrm{len}}, w_{\mathrm{yaw}}, w_{\mathrm{lat}}$ are omitted 
here as they are tuned per sensor configuration and dataset.
Starting from each priority lane, a connectivity search constructs the set of 
continuously connected lanes in the To direction while excluding geometrically 
inconsistent lanes and duplicates between Tracking and TemporaryLost states.

\textbf{Output Topology Correction.}
For the LCLC of the lane set selected in the previous step, 
inconsistent connections are removed based on endpoint distance and heading alignment, 
and missing connections are added for lane pairs satisfying the conditions.
Competing connections are resolved to retain only the most geometrically
consistent one, suppressing crossing LCLC and spurious connections.

\textbf{Near-Endpoint Smoothing for Connectivity Alignment.}
At lane junctions the local point sequence near the terminal end of $s_u$​ and 
the start of $s_v$​ may not coincide, 
causing positional gaps and directional kinks 
that destabilize downstream topology interpretation and path generation.
For each connected lane pair, a point sequence of fixed length $N$ is extracted
from the terminal side of $s_u$ and from the start side of $s_v$; a
B\'ezier curve of degree $d$ is fitted to these with position and tangent
constraints (C0/C1) imposed at the outer endpoints:
\begin{equation}
B(t) = \sum_{i=0}^{d} \binom{d}{i}(1-t)^{d-i}\,t^{i}\,P_i,\quad t \in [0,1]
\end{equation}
Control points $\{P_i\}$ are determined from the outer-endpoint
position/tangent constraints together with approximation of the inner local
points (e.g., by least squares).
Both lanes' near-endpoint point sequences are replaced by points resampled from
$B(t)$, resolving C0 and C1 discontinuities at the junction.
This processing is applied only to connections within the selected lane set 
that are validated by topology correction; 
connections failing the acceptance conditions are invalidated rather than smoothed.

\subsection{Resource-Constrained Real-Time Design}\label{ssec:realtime_design}

Because per-frame cost grows quadratically with input size, 
real-time operation (10 Hz, 20 ms) is ensured by three principles. 

First, frame-internal preprocessing, gating, and heading-based partitioning 
(Section~\ref{ssec:refinement} and \ref{ssec:association}) reduce the observation–buffer pair space before the dominant cost stage. 
Second, the same-index Fréchet approximation (Section~\ref{ssec:association}) reduces 
per-pair cost from $O(n^2)$ to $O(n)$. 
Third, each functional block has a wall-clock budget; on overrun, 
low-priority matching and state updates are skipped 
and the previous-frame output is retained to avoid starving the downstream planner.

Runtime measurements (Section~\ref{ssec:runtime}) show that the average remains well within the budget, 
while at peak load the worst-case reaches the limit, 
where the fallback mechanism ensures cycle-time guarantees.

\section{Experiments}

\subsection{Evaluation Design}

\subsubsection{Evaluation Framework}

To isolate improvements in lane structure quality, we decompose it into four components:
$\mathcal{Q}_{\text{topology}} = (\mathcal{D}, \mathcal{G}, \mathcal{A}, \mathcal{T})$,
where $\mathcal{D}$ denotes detection accuracy, $\mathcal{G}$ geometry consistency,
$\mathcal{A}$ attribute correctness, and $\mathcal{T}$ topology validity.

In addition, we evaluate across-frame stability of connectivity ($\mathcal{T}$) 
and labels ($\mathcal{A}$) using temporal metrics (Section~\ref{ssec:metrics}), 
since per-frame metrics such as $F1$ cannot capture temporal consistency.

\subsubsection{Dataset and Matching}\label{ssec:matching}

Evaluation uses the validation dataset of the perception model (16 sequences with ground truth), 
where WASABI post-processes the inference results. 
We use an internal dataset rather than public benchmarks 
(e.g., OpenLane-V2~\cite{wang2023openlanev2}) due to licensing constraints, 
so absolute values are not directly comparable and comparison with external baselines is left to future work.
Because GT, inference, and post-processing outputs have different sampling densities, 
all curves are resampled to 10 points, and similarity is evaluated using the discrete Fréchet distance $d_F$.

The distance between each predicted lane $\tilde{P}_i$ and GT lane $\tilde{G}_j$ is defined as:
\begin{equation}
D_{ij} = d_F(\tilde{P}_i, \tilde{G}_j)
\end{equation}
Pairs with $D_{ij} < \tau$ ($\tau = 3.0\,\mathrm{m}$) are treated as candidates, 
and one-to-one greedy matching in ascending $D_{ij}$ yields $\hat{\mathcal{M}}$. 
We define $\mathrm{TP} = |\hat{\mathcal{M}}|$, 
$\mathrm{FP} = |\mathcal{P}| - \mathrm{TP}$, 
and $\mathrm{FN} = |\mathcal{G}| - \mathrm{TP}$.

\subsubsection{Metrics}\label{ssec:metrics}

$\mathcal{D}$ is evaluated by Precision/Recall/F1.
$\mathcal{G}$ by lateral deviation from GT lanes, defined as:
\begin{equation}
\varepsilon_{\mathrm{geo}} = \frac{1}{N} \sum_{i=1}^{N} d_{\perp}(p_i, L_{\mathrm{GT}})
\end{equation}
where $d_{\perp}(p_i, L_{\mathrm{GT}}) = \min_{s \in L_{\mathrm{GT}}} \|p_i - s\|$.
$\mathcal{A}$ is evaluated by attribute accuracy on TP-matched lanes, 
and $\mathcal{T}$ by Precision/Recall/F1 over LCLC between matched lane pairs.
LCLC evaluation is restricted to connections whose endpoint lane segments are both TP-matched to GT, 
ensuring fair comparison independent of detection-stage differences.

\textbf{Temporal Stability Metrics.}
To assess temporal stability independently of per-frame accuracy, we define two GT-anchored metrics. 
For each consecutive frame pair $(t, t{+}1)$, predictions are ego-motion compensated and 
independently TP-matched to GT at frame $t{+}1$ using the same Fr\'echet-based matcher and 
threshold $\tau = 3.0\,\mathrm{m}$ as in Section~\ref{ssec:matching}. 
Only GT lanes that obtain a TP match in both frames are used,
yielding a common population for fair comparison; the WASABI
Track ID is not used for this association.

(i) \textbf{LCLC Toggle Rate} measures the fraction of connections between such lane pairs 
whose state flips between TRUE and FALSE across frames. 
The baseline LCLC scores are binarized using the same threshold as WASABI.

(ii) \textbf{Boundary-Label Flicker Rate} measures the frequency of changes in boundary labels 
(left/right $\times$ type/color) for the same lane, averaged across the four fields.

Both metrics are computed per sequence and averaged over all sequences; 
lower values indicate greater temporal stability.

\subsection{Main Results}

WASABI post-processing results are compared with baseline perception model outputs 
across four axes: 
Detection (Table~\ref{tab:lane_detection_eval}), 
Geometry (Table~\ref{tab:lane_metric_eval}), 
Attribute (Table~\ref{tab:lane_label_eval}), 
and LCLC (Table~\ref{tab:lclc_detection_eval}). 
Consistent improvements are observed across all metrics; 
details are given below.

\textbf{(A) Lane segment detection accuracy ($Q_D$)}

The primary driver of improvement is FP reduction
(Table~\ref{tab:lane_detection_eval}).
FP count decreases from 328,459 to 247,639, a reduction of approximately
24.6\%, and Precision improves from 0.367 to 0.465 (+26.8\% relative).
Recall also improves from 0.326 to 0.369, demonstrating that false-positive
suppression is achieved without sacrificing detection coverage.
Detection F1 improves from 0.345 to 0.412 (+0.067, +19.2\%).

\begin{table}[!t]
  \centering
  \caption{Lane detection evaluation.}
  \label{tab:lane_detection_eval}
  \footnotesize
  \setlength{\tabcolsep}{3pt}
  \resizebox{\columnwidth}{!}{%
    \begin{tabular}{lcccccc}
      \toprule
      & TP & FP & FN & Precision & Recall & F1-score \\
      \midrule
      Perception model
        & 190,169 & 328,459 & 392,276 & 0.367 & 0.326 & 0.345 \\
      WASABI
        & \textbf{215,167} & \textbf{247,639} & \textbf{367,278}
        & \textbf{0.465} & \textbf{0.369} & \textbf{0.412} \\
      \bottomrule
    \end{tabular}%
  }
\end{table}

\textbf{(B) Geometry consistency ($Q_G$)}

Centerline lateral error improves substantially from 2.50~m to 0.95~m,
and boundary lateral error from 3.21~m to 1.06~m (Table~\ref{tab:lane_metric_eval}).
The primary driver of this improvement is that preprocessing and output lane
selection remove low-geometry-quality duplicate and false-positive lanes,
making the lane set delivered to the downstream planner a less noisy subset
of the raw perception output.
Note that the near-endpoint B\'ezier smoothing (Section~III-C) targets
positional and directional continuity at junctions, 
and its contribution to this metric (lateral error to GT lanes) is limited.

\begin{table}[!t]
  \centering
  \caption{Lane geometry metrics evaluation.}
  \label{tab:lane_metric_eval}
  \small
  \setlength{\tabcolsep}{4pt}
  \renewcommand{\arraystretch}{1.12}
  \resizebox{\columnwidth}{!}{%
  \begin{tabular}{lcc}
    \toprule
    & CL Lateral Error (m) & Lane Mark Lateral Error (m) \\
    \midrule
    Perception model & 2.4961 & 3.2129 \\
    WASABI & \textbf{0.9512} & \textbf{1.0647} \\
    \bottomrule
  \end{tabular}%
  }
\end{table}

\textbf{(C) Attribute correctness ($Q_A$)}

All four boundary label categories improve marginally over the baseline
(Table~\ref{tab:lane_label_eval}).
Temporal smoothing via the tracking buffer and enforced attribute consistency
through matching suppress inter-frame label flicker, though absolute accuracy
remains in the 0.74--0.83 range; label stabilization is not the primary effect
of this method.

\begin{table}[!t]
  \centering
  \caption{Boundary line label evaluation.}
  \label{tab:lane_label_eval}
  \small
  \setlength{\tabcolsep}{4pt}
  \renewcommand{\arraystretch}{1.10}
  \begin{tabular}{lcccc}
    \toprule
    & left\_type & right\_type & left\_color & right\_color \\
    \midrule
    Perception model & 0.738 & 0.782 & 0.740 & 0.824 \\
    WASABI & \textbf{0.755} & \textbf{0.799} & \textbf{0.757} & \textbf{0.831} \\
    \bottomrule
  \end{tabular}
\end{table}

\textbf{(D) Topology detection accuracy ($Q_T$)}

LCLC F1 shows the largest improvement (0.834 $\to$ 0.948, +0.114, +13.6\%)
(Table~\ref{tab:lclc_detection_eval}).
FP edges decrease by $-70.1\%$, and FN edges by $-50.6\%$.

\begin{table}[!t]
  \centering
  \caption{LCLC detection evaluation.}
  \label{tab:lclc_detection_eval}
  \small
  \setlength{\tabcolsep}{4pt}
  \renewcommand{\arraystretch}{1.10}
  \resizebox{\columnwidth}{!}{%
  \begin{tabular}{lcccccc}
    \toprule
    & TP & FP & FN & Precision & Recall & F1-score \\
    \midrule
    Perception model & 44,331 & 7,660 & 9,924 & 0.853 & 0.817 & 0.834 \\
    WASABI & \textbf{65,651} & \textbf{2,291} & \textbf{4,906}
              & \textbf{0.966} & \textbf{0.930} & \textbf{0.948} \\
    \bottomrule
  \end{tabular}%
  }
\end{table}

\begin{table}[t]
\centering
\caption{Temporal stability evaluation (GT-anchored, 16 sequences; lower is
more stable).}
\label{tab:temporal}
\begin{tabular}{lcc}
\toprule
 & LCLC Toggle Rate & Boundary-Label Flicker Rate \\
\midrule
Perception model & 0.00365 & 0.0389 \\
WASABI           & 0.00134 & 0.0271 \\
\midrule
$\Delta$ (relative) & $-63.3\%$ & $-30.2\%$ \\
\bottomrule
\end{tabular}
\end{table}

\textbf{(E) Across-frame temporal stability}

The GT-anchored temporal metrics confirm stabilization across frames that 
per-frame F1 cannot capture (Table~\ref{tab:temporal}).
The LCLC toggle rate decreases by 63.3\% (0.00365 $\to$ 0.00134), and 
the boundary-label flicker rate by 30.2\% (0.0389 $\to$ 0.0271). 
Flicker reduction is observed across all sixteen sequences 
and all four label fields (per-field reductions of −43.8\% $\to$ −49.2\%), 
while the toggle rate improves in 14 of 16 sequences.
These results complement the per-frame LCLC F1 (Table~\ref{tab:lclc_detection_eval}), 
indicating that recovered connections are not only more accurate 
per frame but also more temporally consistent.

\subsection{Ablation Study}\label{ssec:ablation}

Table~\ref{tab:ablation} reports five variants analyzing the three main contributions.

\textbf{w/o Tracking} disables temporal integration (history accumulation, state management, and the $W{=}10$ majority vote).

\textbf{w/o Refinement} disables both refinement stages; 
\textbf{w/o Preprocessing} and \textbf{w/o Output Sel./Corr.} further isolate the input- and output-side contributions.

\textbf{Exact Fr\'echet} replaces the same-index approximation.

\begin{table}[!t]
  \centering
  \caption{Ablation results for WASABI components. The last two columns are
  GT-anchored temporal metrics (Section~\ref{ssec:metrics}; lower is more
  stable). Temporal metrics are measured for the Tracking ablation only.}
  \label{tab:ablation}
  \small
  \setlength{\tabcolsep}{3pt}
  \begin{adjustbox}{max width=\columnwidth}
    \begin{tabular}{lcccccc}
      \toprule
      Variant & Det.\ F1 & CL error (m) & Attr.\ Acc. & LCLC F1
              & Toggle$_{\text{gt}}$ & Flicker$_{\text{gt}}$ \\
      \midrule
      Full WASABI            & \textbf{0.412} & \textbf{0.95} & $\approx \mathbf{0.79}$
                               & \textbf{0.948} & \textbf{0.00134} & \textbf{0.0271} \\
      (a) w/o Tracking       & 0.422 & 0.95 & $\approx 0.79$ & 0.942 & 0.00142 & 0.0278 \\
      (b) w/o Refinement (c+d) & 0.391 & 0.93 & $\approx 0.78$ & 0.938 & --- & --- \\
      (c) w/o Preprocessing  & 0.410 & 0.95 & $\approx 0.79$ & 0.947 & --- & --- \\
      (d) w/o Output Sel./Corr. & 0.403 & 0.95 & $\approx 0.78$ & 0.942 & --- & --- \\
      (e) Exact Fr\'echet    & 0.412 & 0.95 & $\approx 0.79$ & 0.948 & --- & --- \\
      \bottomrule
    \end{tabular}
  \end{adjustbox}
  \par\smallskip
\end{table}

\vspace{0.3em}

\textbf{Tracking contribution.}
Detection F1 nominally increases without Tracking (0.412$\to$0.422) because 
Full WASABI withholds newly observed lanes until $N_{TrackStMove}{=}3$ consecutive associations. 
Disabling this gate emits ${\sim}7{,}000$ additional lanes (TP $215{,}167\to222{,}225$, 
FN drops by the same amount, FP remains nearly unchanged at $247{,}639\to247{,}668$). 

Frame-independent F1 therefore conflates output gating with recall and 
is not suitable for assessing the Tracking contribution, 
whose role is to produce temporally consistent outputs for the downstream planner.

LCLC F1 drops slightly ($0.948\to0.942$), as output-side correction partially compensates 
for the absence of temporal integration in this variant.

The GT-anchored temporal metrics in Table~\ref{tab:ablation} show consistent degradation: 
toggle rate increases ($0.00134 \to 0.00142$) and 
flicker rate increases ($0.0271 \to 0.0278$). 
The magnitude is modest because output-side correction absorbs part of the temporal role. 
Thus, the Tracking contribution to across-frame stability is measurable 
but partially redundant with Refinement in this dataset.

\vspace{0.3em}

\textbf{Refinement contribution.}
Disabling both refinement stages increases FP from $247{,}639$ to $356{,}097$, 
exceeding even the raw baseline ($328{,}459$), 
indicating that tracking without structuring re-emits duplicates and spurious connections over time.

Individual ablations show only small F1 drops ($-0.002$ and $-0.009$ 
for input- and output-side, respectively), 
whereas the joint removal results in a larger drop ($0.021$). 
FP exhibits the same superadditive pattern (individual increases of $+3{,}378$ and $+61{,}378$ vs.\ a combined increase of $+108{,}458$).

These results indicate that input- and output-side stages act complementarily, 
each compensating for the absence of the other, 
with output-side correction being the dominant FP suppressor.

The w/o Refinement variant effectively serves as a tracking-only baseline, 
retaining association, state management, and LCLC history 
while removing both refinement stages. 
Its detection FP exceeds even the raw perception model, 
and its LCLC F1 (0.938) falls below the full pipeline (0.948).

This confirms that tracking alone does not yield the observed gains; 
WASABI’s improvements arise from the combination of temporal association and noise-robust refinement.

\vspace{0.3em}

\textbf{Fréchet approximation validity.}
Exact Fr\'echet yields results consistent with Full WASABI across all evaluation metrics, 
indicating that the same-index approximation introduces no measurable accuracy loss. 
Geometry error (0.93--0.95~m) and attribute accuracy ($\approx$0.78--0.79) 
remain effectively unchanged across variants, 
confirming that the approximation preserves both geometric fidelity and semantic consistency.

\subsection{Runtime and Scalability}\label{ssec:runtime}

WASABI satisfies the onboard cycle-time constraint, 
as confirmed on the target hardware (a single Arm Cortex-A78 CPU core). 
To stress the system under heavy load, runtime is evaluated 
on a large U.S. intersection scene distinct from the validation set 
(Sections~\ref{ssec:matching}--\ref{ssec:ablation}), 
where per-frame input and tracked-lane counts reach up to 126 and 135, respectively. 
In this setting, cost computation and data association 
exceed the cycle-time budget without the real-time optimizations. 
With the heading-partitioned matching (Section~\ref{ssec:segment_tracking}), 
this stage is brought within budget (Table~\ref{tab:runtime}: 0.625 ms average, 1.298 ms worst). 
Table~\ref{tab:runtime} reports per-block runtimes: 
the average total is 7.457 ms, about one third of the 20 ms budget, 
with Track State Update as the dominant stage. 
At peak load, the worst-case total reaches 20.558 ms, 
where the per-block wall-clock budget and early-termination fallback (Section~\ref{ssec:realtime_design}) engage, 
retaining the previous-frame output and preventing starvation of the downstream planner.

\begin{table}[!t]
  \centering
  \caption{Per-functional-block runtime on a large intersection scene (distinct from the validation set; single Arm Cortex-A78 core).}
  \label{tab:runtime}
  \small
  \setlength{\tabcolsep}{3pt}
  \renewcommand{\arraystretch}{1.08}
  \begin{tabular}{p{4.2cm}rr}
    \toprule
    Functional block & avg (ms) & worst (ms)\\
    \midrule
    Lane Segment Preprocessing & 1.520 & 4.775 \\
    Input Coordinate Transform & 0.283 & 0.603 \\
    Cost Matrix Construction \& Data Association & 0.625 & 1.298 \\
    Track State Update & 2.583 & 7.584 \\
    Output Coordinate Transform & 0.387 & 0.661 \\
    Output Lane Selection & 1.246 & 4.571 \\
    Node-Edge Graph Construction & 0.809 & 2.098 \\
    \midrule
    Sum & 7.457 & 20.558 \\
    \bottomrule
  \end{tabular}
  \par\smallskip
\end{table}

\section{Conclusion}

This paper presents WASABI, a real-time post-processing pipeline that 
jointly stabilizes lane segments and their topological connectivity.

As shown in Section~IV, Segment Tracking with Connectivity provides the largest gain 
in LCLC F1 by leveraging connection history to recover structure that 
is unstable in single-frame predictions. 
Detection improvements are primarily driven by FP reduction, 
suppressing spurious lane candidates that degrade downstream performance. 
GT-anchored temporal metrics further demonstrate that stabilization 
is achieved not only in per-frame accuracy but also across frames 
(LCLC toggle rate $-63.3\%$, boundary-label flicker rate $-30.2\%$).

Limitations remain in absolute detection performance (Precision and Recall below 0.5) and 
attribute accuracy (0.74--0.83), as well as in the quadratic growth of computational cost 
with input size. 

Future work includes improving robustness to difficult scenes, 
improving computational efficiency and mitigating the quadratic scaling with input-lane count, 
and improving endpoint stability and closed-loop evaluation with the behavior planner.

\printbibliography

\end{document}